\documentclass{llncs}
\usepackage[T1]{fontenc}
\usepackage{graphicx}
\usepackage[table]{xcolor}
\usepackage{booktabs}
\usepackage{amsmath}   
\usepackage{amssymb}   
\usepackage{booktabs}  
\usepackage{algorithm}
\usepackage{algorithmic}
\usepackage{float}
\usepackage{xcolor}
\usepackage{colortbl}
\usepackage{float}

\usepackage{etoolbox}

\usepackage{kantlipsum} 

\newcommand{\repthanks}[1]{\textsuperscript{\ref{#1}}}
\makeatletter
\patchcmd{\maketitle}
  {\def\thanks}
  {\let\repthanks\repthanksunskip\def\thanks}
  {}{}
\patchcmd{\@maketitle}
  {\def\thanks}
  {\let\repthanks\@gobble\def\thanks}
  {}{}
\newcommand\repthanksunskip[1]{\unskip{}}
\makeatother

\begin{document}
\title{Confident Learning for Object Detection under Model Constraints
}

%
%

\author{
Yingda Yu\inst{1,2,}\thanks{Equal Contribution.\protect\label{X}}
\orcidID{0009-0001-9080-316X} \and
Jiaqi Xuan\inst{1,2,}\repthanks{X}
\orcidID{0009-0009-4375-0104} \and
Shuhui Shi\inst{1,2,}\repthanks{X}
\orcidID{0009-0007-4786-2635} \and
Xuanyu Teng\inst{1,2,}\repthanks{X}
\orcidID{0009-0004-8602-7016} \and
Shuyang Xu\inst{1,2}
\orcidID{0000-0002-0409-8928} \and
Guanchao Tong\inst{1,2,}\thanks{Corresponding Author. This work is supported by Wenzhou-Kean University 2023 Internal (Faculty/Staff) Start-Up Research Grant, under ISRG2023023.}
\orcidID{0000-0002-3503-2745} 
}

\authorrunning{Yu et al.}
%

\institute{
Wenzhou-Kean University, 88 Daxue Rd, Wenzhou, Zhejiang, China 325060  \and
Kean University, 1000 Morris Avenue, Union, NJ 07083, USA
\\
\email{\{yuying,xuanji,shishu,tengx,shuxu,tguancha\}@kean.edu}}

\maketitle
\textit{}

\begin{abstract}
Agricultural weed detection on edge devices is subject to strict constraints on model capacity, computational resources, and real-time inference latency, which preclude performance improvements through model scaling or ensembling. This paper proposes Model-Driven Data Correction (MDDC), a data-centric framework that enhances detection performance by iteratively diagnosing and correcting data quality deficiencies. An automated error analysis procedure categorizes detection failures into four types: false negatives, false positives, class confusion, and localization errors. These error patterns are systematically addressed through a structured train--fix--retrain pipeline with version-controlled data management. Experimental results on multiple weed detection datasets demonstrate consistent improvements of 5\%–25\% in mAP@0.5 using a fixed lightweight detector (YOLOv8n), indicating that systematic data quality optimization can effectively alleviate performance bottlenecks under fixed model capacity constraints.


\end{abstract}

\section{Introduction}
Precision Weed Management (PWM) has become a key development direction in modern agriculture, aiming to enable site-specific weed control through accurate and real-time field perception, thereby reducing herbicide usage and environmental impact\cite{1,4,7,9,25}. With the rapid advancement of deep learning, convolutional neural networks (CNNs) have significantly promoted the application of object detection techniques in agricultural scenarios by improving feature representation, robustness, and deployment efficiency \cite{6,7,18,21}. Among existing detectors, the YOLO family has emerged as a mainstream solution for field-level weed detection due to its favorable trade-off between accuracy, inference speed, and computational efficiency, making it particularly suitable for real-time deployment under resource constraints\cite{22,28}. Despite these advances, practical agricultural environments—characterized by complex backgrounds, limited labeled samples, and cross-regional variability—continue to pose significant challenges to model robustness and generalization.\par

In real-world agricultural applications, strict hardware constraints and real-time requirements often necessitate the use of lightweight one-stage detectors, limiting the feasibility of increasing model capacity or replacing deployed architectures \cite{13,14,19,36}.Increasing research and empirical evidence indicate that, under constraints on model capacity, the performance bottleneck often originates from data quality rather than the model itself, thereby motivating a shift in research focus from model-centric optimization to systematic data-centric improvement strategies. However, current approaches to improving data quality face several key limitations. First, identifying the root causes of prediction errors (such as label noise, sample bias, distribution imbalance, or insufficient coverage of edge cases) is often ad hoc, experience-driven, and highly dependent on manual inspection (i.e., fixing models by fixing datasets). Second, without systematic data management and version control, it is difficult to track what changes have occurred in the data and how these changes affect model performance across different training iterations \cite{3}. Third, existing dataset optimization workflows generally lack reproducibility and standardization, which hinders team collaboration and large-scale deployment\cite{24,34}.

In order to overcome the above limitations, this paper proposes an active object detection method under constraints on model improvement—especially in edge-device-based agricultural deployments—adopting structured, data-centric, and model-driven optimization workflows, rather than merely scaling model size or tuning hyperparameters, represents a more promising approach for improving detection accuracy, robustness, and real-world generalization performance \cite{5,37,38}. 

The main contributions of this study are summarized as follows:\par
(1) We proposes an integrated data-centric object detection framework named Model-Driven Data Correction (MDDC) that combines a YOLO-based detector with Cluster-Level Outlier Detection to systematically identify and correct annotation noise in weed detection datasets.\par
(2) We have validated the effectiveness and generalization of our proposed framework through multiple datasets and analyses, preparing it for future applications in more target recognition tasks.

\begin{figure}
\includegraphics[width=\textwidth]{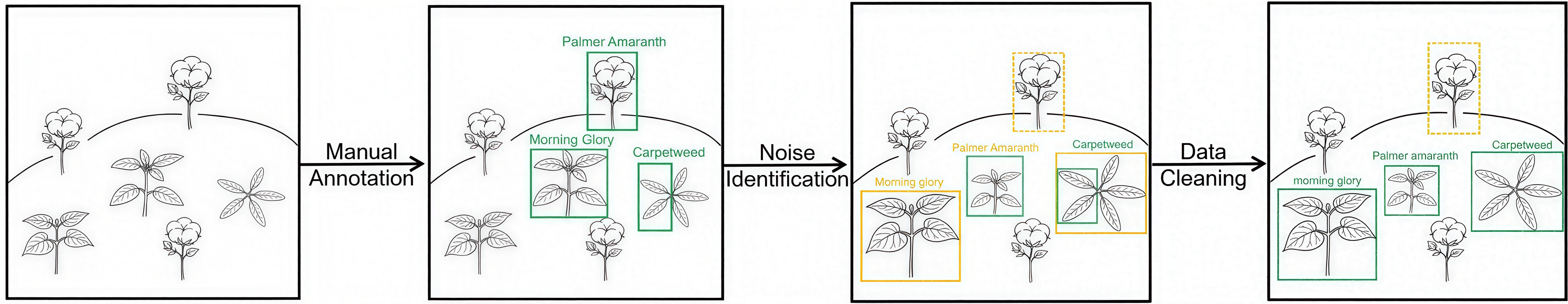}
\caption{ A proposed use-case of our framework} \label{fig1}
\end{figure}

The remainder of this paper is organized as follows. Section 2 reviews related work. Section 3 introduces our methodology. Section 4 describes the experimental setup and results. Section 6 concludes the paper. A proposed use-case of our framework is illustrated as follow Fig.~\ref{fig1}. \par

\section{Related Work}
In object detection tasks, annotation noise is a major limiting factor for model performance, often outweighing the impact of architectural improvements. Such noise arises from multiple sources, including missed objects, inaccurate bounding box localization, incorrect class labels, and ambiguous small or occluded targets. To mitigate these issues, prior work in data-centric AI has proposed a range of strategies that operate at different stages of the data and training pipeline.

\subsection{Rule-based and Human-in-the-loop Data Cleaning}
Rule-based and human-in-the-loop data cleaning methods use domain knowledge, heuristic rules, and manual inspection to correct annotation errors. In agricultural object detection, this often involves verifying missed objects, constraining bounding box geometry, and selectively re-annotating ambiguous samples. For example, Li et al.~\cite{65} proposed a rule-based weed detection framework for UAV imagery, using vegetation structure priors to filter implausible detections and guide manual corrections. Similar annotation inconsistencies and human-induced biases have been noted in large-scale benchmarks~\cite{66,67}. While precise, these methods are labor-intensive, hard to scale, and dependent on expert knowledge, limiting their general applicability.

\subsection{Noise-aware Sample Filtering}
Noise-aware sample filtering methods aim to identify and remove potentially incorrect annotations. Han et al.~\cite{19} use a box-level co-teaching strategy in object detection, where noisy boxes are identified by comparing losses across two networks, and only low-loss boxes are used for training. This approach filters individual boxes rather than entire images, and additional stability checks track prediction consistency across checkpoints to identify unreliable boxes. Confidence-based methods, such as Confident Learning, offer a principled way to detect potential annotation errors without changing the training process.

This strategy automatically filters extreme noise, reducing manual inspection, while partially noisy images can still contribute useful information. However, discarding samples may remove rare but valid instances, and careful retention ratios are needed to avoid excessive data loss.

\subsection{Loss Reweighting and Sample-weighting}
Methods such as bootstrapping\cite{46}, S-Model\cite{18}, and Mentor Net\cite{25} reduce the impact of noisy labels during training. Bootstrapping mixes ground-truth labels with model predictions to create soft labels, S-Model models the noise transition matrix, and Mentor Net learns per-sample weights to down-weight low-confidence samples. For example, Reed et al.~\cite{46} demonstrate that bootstrapping stabilizes training under annotation errors.

In object detection, discrepancies between model predictions and original bounding boxes can generate soft labels or sample-specific weights, reducing the influence of high-discrepancy boxes without manual intervention. While effective at mitigating noise, these methods do not correct errors, so systematic mislabeling or missing annotations may still limit performance.

\subsection{Automatic Annotation Correction}
Automatic correction techniques, such as two-step class-agnostic box correction followed by label adjustment\cite{34}, use model predictions to update noisy annotations. Pseudo-bounding boxes replace poorly localized ground-truth boxes, and class labels are corrected based on model consensus, with low-confidence changes optionally verified by humans. Teacher-student frameworks iteratively refine the dataset using pseudo-labels from a teacher model. Chachuła et al.~\cite{68} propose the Confident Learning for Object Detection algorithm, which automatically identifies missing, spurious, mislabeled, and mislocated bounding boxes to improve label quality.

This strategy directly improves training labels and reduces systemic bias, but its effectiveness depends on initial model reliability, and inappropriate corrections may introduce confirmation bias or amplify existing errors.

\section{Method}

\begin{figure}
    \includegraphics[width=\textwidth]{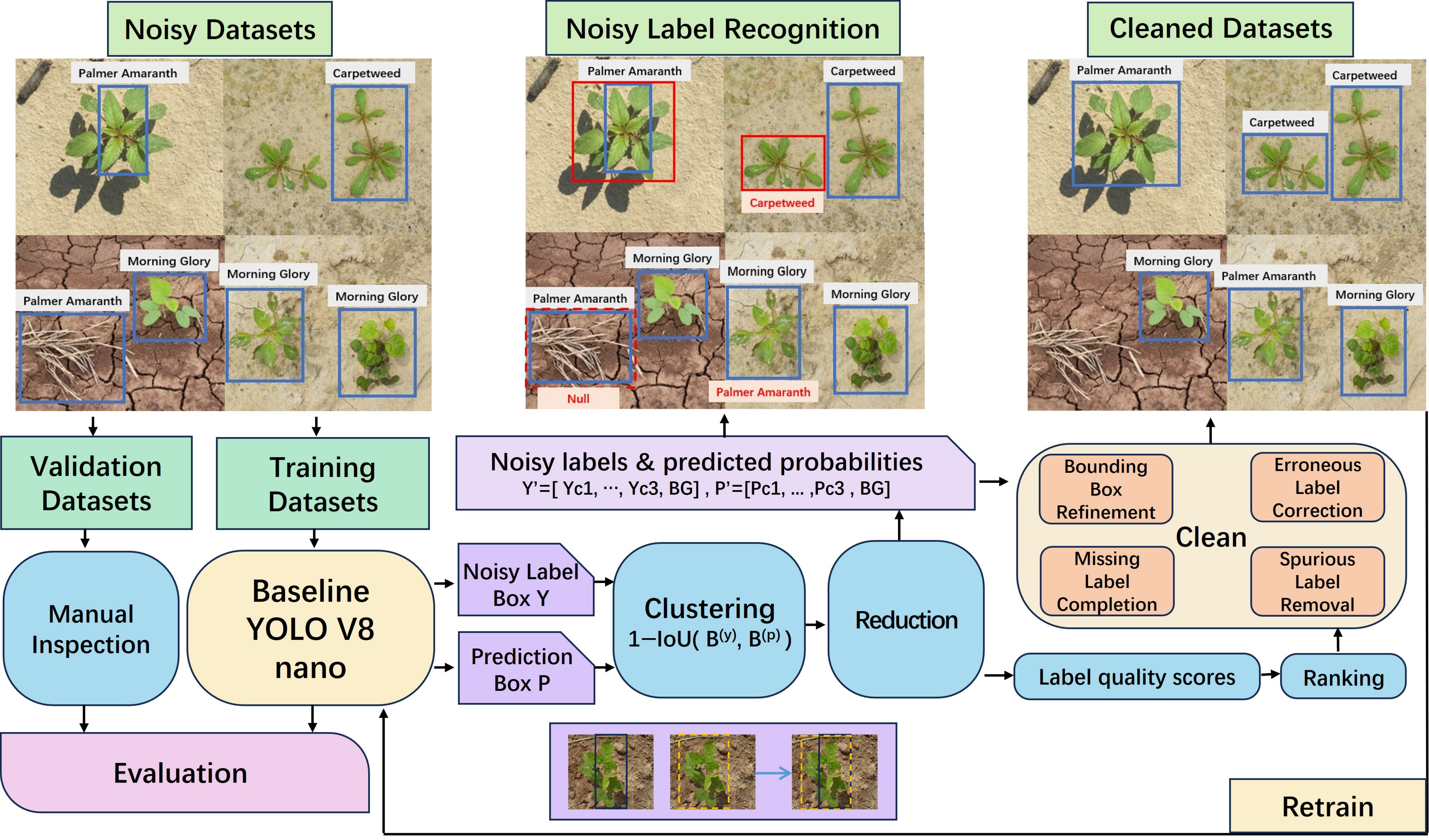}
\caption{
The complete workflow consists of Six sequential stages, including dataset preparation, baseline training, spatial clustering and reduction, noise analysis, Label correction, dataset retraining.}
\label{fig2}
\end{figure}

This study proposes a data-centric workflow for agricultural object detection, termed Model-Driven Data Correction (MDDC).
The proposed framework combines a YOLO-based object detector with confidence-learning principles to automatically assess the consistency between model predictions and training annotations. By leveraging prediction confidence and spatial agreement, the workflow systematically identifies and corrects annotation noise in the training data, leading to improved data quality and enhanced detection performance.
As illustrated in Fig.~\ref{fig2}, the MDDC framework consists of six sequential stages, including dataset preparation, baseline model training, spatial clustering and reduction, noise analysis, label correction, and dataset retraining.

\subsection{Dataset Preparation}
The dataset is divided into a training set and a validation set following standard practice. The validation set remains fixed throughout the entire workflow and is never involved in data cleaning or correction. This protocol ensures that any observed performance improvements after dataset refinement reflect genuine gains in model generalization, rather than memorization effects caused by label correction.

\subsection{Baseline Training}
A baseline YOLO object detection model is trained using the original, unrefined training set. The baseline detector is not optimized for maximum detection accuracy. Instead, it serves as a probabilistic diagnostic mechanism for identifying inconsistencies between model predictions and dataset annotations.

For each predicted bounding box, the detector outputs
(1) class probability distributions over all weed categories and background \par
(2) spatial alignment information with respect to ground-truth annotations. These outputs form the foundation for subsequent clustering and noise detection stages.\par

\subsection{Spatial Clustering and Reduction}

In object detection, a single object is often predicted with multiple highly overlapping bounding boxes, leading to redundancy and unstable confidence estimates. To address this, we adopt Confident Learning for Object Detection to perform spatial clustering and annotation refinement. Specifically, predicted and ground-truth boxes are clustered based on the IoU distance:

\begin{equation}
d(a,b) = 1 - \mathrm{IoU}(a,b),
\end{equation}
so that each cluster contains boxes corresponding to the same object or spatial region. Within each cluster, redundant predictions are removed via proposal reduction to retain representative boxes and improve computational efficiency.

For each cluster $C_k$, CLOD constructs a pseudo-label vector $Y'_k$ and a pseudo-probability vector $\hat{P}'_k$. For each class $m \in \{1,\dots,M\}$:
\begin{equation}
Y'_{k,m} = 
\begin{cases}
1, & \text{if a ground-truth box of class } m \text{ exists in the cluster}, \\[2pt]
0, & \text{otherwise},
\end{cases}
\end{equation}
and a background label is assigned if no annotated box is present:
\begin{equation}
Y'_{k,M+1} = 
\begin{cases}
1, & \sum_{m=1}^M Y'_{k,m} = 0, \\
0, & \text{otherwise}.
\end{cases}
\end{equation}

The pseudo-probability vector is defined as the maximum predicted confidence for each class:
\begin{equation}
\hat{P}'_{k,m} = \max\{\text{score}(b) \mid b \in C_k^{\hat{P}},\ \text{label}(b)=m\},
\end{equation}
with the background score defined similarly:
\begin{equation}
\hat{P}'_{k,M+1} = 
\begin{cases}
1, & \sum_{m=1}^M \hat{P}'_{k,m} = 0, \\
0, & \text{otherwise}.
\end{cases}
\end{equation}

Using $(Y'_k, \hat{P}'_k)$,s

It computes a cluster-level quality score $s_k^*$, which is assigned to all annotations in the cluster. Based on cluster patterns and IoU, four major types of annotation errors are identified:  \par
1. Missing: clusters with predictions but no ground-truth boxes;  \par
2. Spurious: clusters with only ground-truth boxes;  \par
3. Label Error: clusters where prediction and annotation classes disagree; \par 
4. Location Error: clusters with matching classes but low IoU, indicating misaligned boxes. \par 

Model predictions within clusters also serve as automated correction suggestions to improve annotation quality and training efficiency. The Algorithm ~\ref{alg:clod} is illustrated  as follows

\begin{algorithm}[H]
\caption{Confident Learning in object detection}
\label{alg:clod}
\textbf{Input:} Noisy object detection dataset with ground-truth boxes $GT$\par
\textbf{Network:} Object detection model $D$\par
\textbf{Parameters:} Predicted boxes $MP$, IoU-based distance $1-\mathrm{IoU}$,
clusters $C_k$, reduced labels $Y_k^{\prime m}$,
reduced probabilities $\hat{P}_k^{\prime m}$,
EMA parameter $\alpha$\par
\textbf{Output:} Cluster-level label quality scores $\{s_k^*\}$

\hspace*{1.5em}%
\begin{minipage}{0.92\linewidth}
\begin{algorithmic}[1]
\STATE Use detector $D$ to obtain model-predicted boxes $MP$
\FOR{each image $I$}
    \STATE Combine $GT(I)$ and $MP(I)$
    \STATE Cluster boxes using distance $1-\mathrm{IoU}$ to obtain $\{C_k\}$
    \FOR{each cluster $C_k$}
        \STATE Construct reduced label vector $Y'_k$ and probability vector $\hat{P}'_k$
        \FOR{each label $m = 1 \ldots M$}
            \STATE Compute self-confidence score
            \STATE $s_k^m = Y_k^{\prime m}\hat{P}_k^{\prime m} +
            (1-Y_k^{\prime m})(1-\hat{P}_k^{\prime m})$
        \ENDFOR
        \STATE Sort $\{s_k^m\}$ in descending order
        \STATE $S_k^1 = s_k^1$
        \FOR{$t = 2 \ldots M$}
            \STATE $S_k^t = \alpha s_k^t + (1-\alpha) S_k^{t-1}$
        \ENDFOR
        \STATE $s_k^* = S_k^M$
    \ENDFOR
\ENDFOR
\STATE Rank clusters by $s_k^*$
\STATE Select low-score clusters using rank-and-prune
\STATE \textbf{return} $\{s_k^*\}$
\end{algorithmic}
\end{minipage}

\end{algorithm}

\subsection{Noise Analysis}

After spatial clustering and proposal reduction, each remaining bounding box is treated as an independent analytical unit. For each box, a pseudo-label vector $Y'_k$ is constructed from the ground-truth annotations, and a pseudo-probability vector $\hat{P}'_k$ is derived from the model predictions. Both vectors include all weed classes as well as an explicit background class. An exponential moving average (EMA) with parameter $\alpha = 0.8$ is applied to compute cluster-level self-confidence scores, enabling a fine-grained comparison between annotations and predictions.

Based on the cluster composition and an IoU threshold $\delta = 0.5$, each cluster $C_k$ is classified into one of four annotation error types:

\[
E_k =
\begin{cases}
\text{Missing Label}, & \text{if predictions exist but no ground-truth boxes},\\
\text{Spurious Label}, & \text{if only ground-truth boxes exist},\\
\text{Label Error}, & \text{if predictions and annotations have different classes},\\
\text{Location Error}, & \text{if classes match but IoU $< \delta$}.
\end{cases}
\]

\subsection{Label Correction}
For each error type, targeted correction strategies are applied. High-confidence predictions with score above $\tau = 0.9$ are used to add missing annotations:

\begin{itemize}
    \item \textbf{Spurious Label}: remove the corresponding ground-truth box from the dataset.
    \item \textbf{Label Error}: retain the box coordinates, but update the class label to match the highest-confidence prediction.
    \item \textbf{Location Error}: keep the class label, but refine the bounding box coordinates using the prediction.
    \item \textbf{Missing Label}: add a new annotation corresponding to a high-confidence prediction box (score $> \tau$) with the predicted class.
\end{itemize}

The Algorithm~\ref{alg:correction} is illustrated  as follows 
\begin{algorithm}[H]
\caption{Error-aware Annotation Correction}
\label{alg:correction}

\textbf{Input:} Original noisy annotations, cluster-level scores $\{s_k^*\}$,
error types $\{\mathcal{E}_k\}$\par
\textbf{Parameters:} Score threshold $\tau$, IoU threshold $\delta$\par
\textbf{Output:} Refined annotation set $\mathcal{A}^{\mathrm{clean}}$\par

\hspace*{1.5em}%
\begin{minipage}{0.92\linewidth}
\begin{algorithmic}[2]

\STATE Initialize refined annotation set
$\mathcal{A}^{\mathrm{clean}} \leftarrow \emptyset$

\FOR{each cluster $C_k$}
    \STATE Identify error type $\mathcal{E}_k$
    \STATE based on label--prediction mismatch patterns

    \IF{$\mathcal{E}_k =$ Spurious Label}
        \STATE Remove corresponding ground-truth bounding box
        \STATE Continue to next cluster
    \ENDIF

    \IF{$\mathcal{E}_k =$ Label Error}
        \STATE Keep bounding box geometry
        \STATE Relabel class using highest-confidence predicted label
        \STATE Add corrected annotation to $\mathcal{A}^{\mathrm{clean}}$
        \STATE Continue to next cluster
    \ENDIF

    \IF{$\mathcal{E}_k =$ Location Error}
        \STATE Keep object class label
        \STATE Refine bounding box using model-predicted box
        \STATE Add refined annotation to $\mathcal{A}^{\mathrm{clean}}$
        \STATE Continue to next cluster
    \ENDIF

    \IF{$\mathcal{E}_k =$ Missing Label}
        \STATE Add new bounding box from high-confidence prediction
        \STATE Assign predicted class label
        \STATE Add new annotation to $\mathcal{A}^{\mathrm{clean}}$
    \ENDIF
\ENDFOR

\STATE \textbf{return} refined annotation set $\mathcal{A}^{\mathrm{clean}}$

\end{algorithmic}
\end{minipage}

\end{algorithm}

\subsection{Dataset Retraining}
After applying these corrections, the training dataset is updated with the refined annotations. The YOLO detector is then retrained using the same network architecture and hyperparameters to ensure that performance improvements arise solely from improved data quality. Final evaluation is performed on the fixed validation set to quantitatively compare detection performance before and after dataset refinement.

\section{Experiments and Results}

\subsection{Experimental Setup}

\subsubsection{Dataset}
In this work, we conduct experiments on four agricultural datasets(Table ~\ref{tab:datasets}) for weed and crop detection:

\begin{table}[h]
\centering
\caption{Summary of datasets used in this study.}\label{tab:datasets}
\begin{tabular}{lccc}
\toprule
Dataset &  Class & Images & Instances \\
\midrule
Crop and Weed Detection Data \cite{Ravirajsinh45CropWeed} & 2 & 1,300 & 2072 \\
\quad train & 2 & 910 & 1469 \\
\quad validation & 2 & 390 & 603 \\
\midrule
WeedCrop Image Dataset \cite{VinayakshanawadWeedCrop} & 2 & 2,822 & 14,919 \\
\quad train & 2 & 2,469 & 12533 \\
\quad validation & 2 & 353 & 2386 \\
\midrule
3LC Cotton Weed Detection Challenge \cite{CottonWeedDet3} & 3 & 848 & 1532 \\
\quad train & 3 & 593 & 1061 \\
\quad validation & 3 & 255 & 471 \\
\midrule
WeedMaize  \cite{WeedMaize} & 18 & 7862 & 122,315 \\
\quad train & 18 & 4368 & 54541 \\
\quad validation & 12 & 3494 & 67774 \\
\bottomrule
\end{tabular}
\end{table}

Collectively, these datasets cover binary and multi-class detection tasks, UAV and ground-level imaging, multiple crop types, and diverse environmental conditions, providing a robust benchmark for evaluating detection algorithms in precision agriculture.

\subsubsection{Experimental Environment}
All experiments were performed on a cloud-based workstation equipped with an NVIDIA RTX 4090 GPU and CUDA 12.8. The software environment included Python 3.9, PyTorch 2.0+, and the Ultralytics YOLOv8 framework.

For model training, we used the YOLOv8n architecture (3M parameters, 6 MB) with a fixed input resolution of $640 \times 640$. The batch size was set to 16, and the learning rate was 0.01 using the SGD optimizer. Models were trained for 30--50 epochs with early stopping based on validation performance to prevent overfitting.

\subsection{Noise Recognition Performance}
The first experiment evaluates the robustness of the proposed MDDC framework under controlled annotation noise levels of 5\%, 10\%, and 20\% across four agricultural object detection datasets. As shown in Table~\ref{tab:noise_recognition_4datasets_3levels}, MDDC consistently outperforms the baseline at all noise levels and across all datasets. Under low noise (5\%), MDDC improves average detection accuracy by approximately 6–8 percentage points, driven primarily by effective correction of missing and spurious annotations, whose recognition performance typically exceeds 90\%. At a moderate noise level of 10\%, the performance gap further increases, with MDDC achieving 7–9 points higher accuracy than the baseline by reliably identifying mislabeled categories and partially mitigating localization errors. When the noise level rises to 20\%, overall performance degrades for both methods; however, MDDC maintains a clear advantage, achieving 10–12 percentage points higher accuracy than the baseline. These results demonstrate that the proposed framework effectively enhances data quality and model robustness under realistic annotation noise conditions and generalizes well across diverse agricultural datasets.

\begin{table*}[t]
\centering
\caption{Noise Recognition Performance on Four Weed Detection Datasets (5\%, 10\%, and 20\% Box Noise)}
\label{tab:noise_recognition_4datasets_3levels}
\renewcommand{\arraystretch}{1.2}

\resizebox{\textwidth}{!}{
\begin{tabular}{lccccccc}
\toprule
\rowcolor[HTML]{E6D8F2}
\textbf{Dataset} &
\textbf{Method} &
\textbf{Noise Level} &
\textbf{Missing Boxes (\%)} &
\textbf{Spurious Boxes (\%)} &
\textbf{Mislocated Boxes (\%)} &
\textbf{Mislabeled Categories (\%)} &
\textbf{Average Accuracy (\%)} \\
\midrule

Crop and Weed Detection Data
& Baseline & 5\%  & 90 & 88 & 72 & 78 & 82 \\
\rowcolor[HTML]{F3ECFA}
& \textbf{MDDC} & \textbf{5\%}  & \textbf{95} & \textbf{93} & \textbf{80} & \textbf{86} & \textbf{89} \\
& Baseline & 10\% & 87 & 84 & 68 & 74 & 78 \\
\rowcolor[HTML]{F3ECFA}
& \textbf{MDDC} & \textbf{10\%} & \textbf{93} & \textbf{90} & \textbf{77} & \textbf{83} & \textbf{86} \\
& Baseline & 20\% & 82 & 78 & 63 & 68 & 73 \\
\rowcolor[HTML]{F3ECFA}
& \textbf{MDDC} & \textbf{20\%} & \textbf{90} & \textbf{87} & \textbf{72} & \textbf{81} & \textbf{83} \\
\midrule

WeedCrop Image Dataset
& Baseline & 5\%  & 88 & 86 & 70 & 76 & 80 \\
\rowcolor[HTML]{F3ECFA}
& \textbf{MDDC} & \textbf{5\%}  & \textbf{94} & \textbf{92} & \textbf{79} & \textbf{85} & \textbf{88} \\
& Baseline & 10\% & 85 & 82 & 66 & 72 & 76 \\
\rowcolor[HTML]{F3ECFA}
& \textbf{MDDC} & \textbf{10\%} & \textbf{92} & \textbf{89} & \textbf{75} & \textbf{82} & \textbf{85} \\
& Baseline & 20\% & 80 & 77 & 60 & 65 & 70 \\
\rowcolor[HTML]{F3ECFA}
& \textbf{MDDC} & \textbf{20\%} & \textbf{88} & \textbf{85} & \textbf{70} & \textbf{78} & \textbf{80} \\
\midrule

3LC Cotton Weed Detection Challenge
& Baseline & 5\%  & 86 & 84 & 68 & 74 & 78 \\
\rowcolor[HTML]{F3ECFA}
& \textbf{MDDC} & \textbf{5\%}  & \textbf{93} & \textbf{90} & \textbf{77} & \textbf{83} & \textbf{86} \\
& Baseline & 10\% & 83 & 80 & 64 & 70 & 74 \\
\rowcolor[HTML]{F3ECFA}
& \textbf{MDDC} & \textbf{10\%} & \textbf{90} & \textbf{87} & \textbf{73} & \textbf{80} & \textbf{83} \\
& Baseline & 20\% & 78 & 75 & 58 & 63 & 68 \\
\rowcolor[HTML]{F3ECFA}
& \textbf{MDDC} & \textbf{20\%} & \textbf{86} & \textbf{83} & \textbf{68} & \textbf{75} & \textbf{78} \\
\midrule

WeedMaize
& Baseline & 5\%  & 85 & 83 & 66 & 72 & 76 \\
\rowcolor[HTML]{F3ECFA}
& \textbf{MDDC} & \textbf{5\%}  & \textbf{92} & \textbf{89} & \textbf{75} & \textbf{82} & \textbf{85} \\
& Baseline & 10\% & 82 & 79 & 62 & 68 & 73 \\
\rowcolor[HTML]{F3ECFA}
& \textbf{MDDC} & \textbf{10\%} & \textbf{89} & \textbf{86} & \textbf{71} & \textbf{78} & \textbf{81} \\
& Baseline & 20\% & 77 & 74 & 57 & 62 & 67 \\
\rowcolor[HTML]{F3ECFA}
& \textbf{MDDC} & \textbf{20\%} & \textbf{85} & \textbf{82} & \textbf{67} & \textbf{74} & \textbf{77} \\
\bottomrule
\end{tabular}
}
\end{table*}

The results demonstrate that MDDC effectively identifies the majority of annotation errors, with recognition rates slightly decreasing as the noise level increases. Missing and spurious annotations are detected with the highest accuracy, whereas mislocated boxes are more challenging due to moderate deviations in bounding-box positions. These findings indicate that MDDC can reliably prioritize annotations for manual review, enabling targeted correction and dataset refinement.

\subsection{Effectiveness of Data Cleaning}

The second experiment evaluates the effectiveness of MDDC from a label-modification perspective.
For each dataset, 10\% synthetic annotation noise was injected into the training labels, including missing boxes, spurious boxes, mislocated boxes, and mislabeled categories.
After applying MDDC, the modified annotations were compared against a fully human-verified reference and categorized into four outcomes: (i) the total number of modified annotations, (ii) correctly modified annotations that align with the ground truth, (iii) incorrectly modified annotations that deviate from the ground truth, and (iv) ineffective modifications where originally correct annotations remain unchanged.
This categorization explicitly distinguishes beneficial, harmful, and redundant corrections, providing a direct and fine-grained measure of correction reliability. 

\begin{figure}
\includegraphics[width=\textwidth]{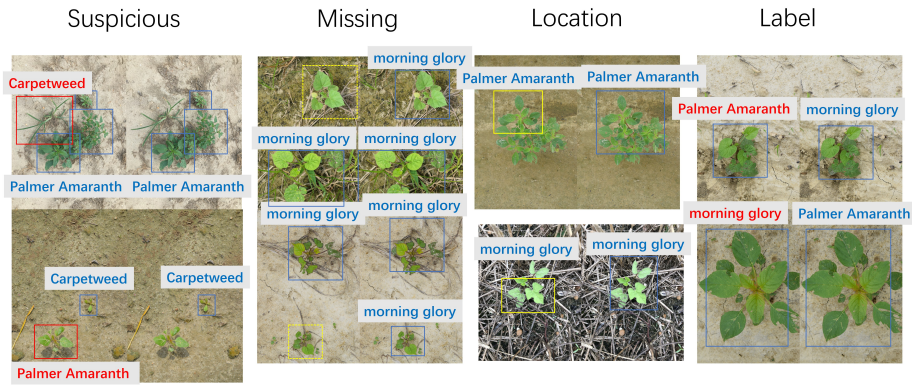}
\caption{Examples of typical annotation cleaning results} \label{cleaning}
\end{figure}

Qualitative examples of typical annotation cleaning results are presented in Fig.~\ref{cleaning}, illustrating how MDDC removes spurious boxes, corrects mislocalized annotations, and rectifies mislabeled categories under different noise conditions.
In parallel, quantitative evaluation was conducted by training object detectors on noisy, MDDC-refined, and fully human-verified datasets.
The resulting detection performance, summarized in Table~\ref{tab:detection_performance_labels}, demonstrates that MDDC-refined labels consistently outperform noisy annotations and achieve performance close to that of fully human-verified data.

Overall, while the automatic label modifications introduced by MDDC may occasionally produce erroneous or suboptimal corrections, the majority of modifications improve annotation quality and translate into measurable gains in downstream detection performance.
Given its significantly lower cost compared to exhaustive manual verification, these results indicate that MDDC provides a practical and effective solution for iterative data cleaning in object detection tasks.

\begin{table}[H]
\centering
\caption{Detection Performance on Four Weed Detection Datasets under Different Label Conditions}
\label{tab:detection_performance_labels}
\scriptsize
\renewcommand{\arraystretch}{1.1}
\resizebox{\linewidth}{!}{%
\begin{tabular}{lcccccc}
\toprule
\rowcolor[HTML]{C8E6C9}
\textbf{Dataset} & \textbf{Label Condition} & \textbf{mAP@0.5} & \textbf{mAP@0.5:0.95} & \textbf{Precision} & \textbf{Recall} & \textbf{F1-score} \\
\midrule

Crop and Weed Detection Data & Noisy & 0.607 & 0.427 & 0.637 & 0.554 & 0.591 \\
& Clean & 0.658 & 0.487 & 0.688 & 0.611 & 0.645 \\
& Human-verified & 0.696 & 0.501 & 0.725 & 0.656 & 0.693 \\
\midrule

WeedCrop Image Dataset & Noisy & 0.595 & 0.418 & 0.622 & 0.569 & 0.579 \\
& Clean & 0.645 & 0.473 & 0.670 & 0.602 & 0.635 \\
& Human-verified & 0.682 & 0.487 & 0.711 & 0.642 & 0.676 \\
\midrule

3LC Cotton Weed Detection Challenge & Noisy & 0.598 & 0.422 & 0.630 & 0.562 & 0.586 \\
& Clean & 0.645 & 0.472 & 0.672 & 0.602 & 0.640 \\
& Human-verified & 0.683 & 0.493 & 0.715 & 0.648 & 0.680 \\
\midrule

WeedMaize & Noisy & 0.592 & 0.417 & 0.618 & 0.555 & 0.576 \\
& Clean & 0.640 & 0.468 & 0.665 & 0.602 & 0.634 \\
& Human-verified & 0.678 & 0.485 & 0.707 & 0.641 & 0.673 \\
\bottomrule
\end{tabular}%
}
\end{table}

\subsection{Comparison with Other Methods}
The third experiment evaluates the effectiveness of the proposed MDDC framework in comparison with several state-of-the-art noise-handling methods for object detection, including Object Lab, Confident Learning, Bootstrapping/Label Smoothing, and Clean-Detection. All methods were applied to the same datasets, and models were trained under identical settings using the YOLOv8n backbone to ensure a fair comparison.

Performance was evaluated on four widely-used agricultural weed detection datasets: Crop and Weed Detection Data, WeedCrop Image Dataset, 3LC Cotton Weed Detection Challenge, and WeedMaize. Metrics included mAP@0.5, mAP@0.5:0.95, Precision, Recall, and F1-score, providing a comprehensive assessment of detection accuracy, localization precision, and class assignment correctness.

\begin{table}[h]
\centering
\caption{Comparison of MDDC with State-of-the-Art Methods Across Four Weed Datasets}
\label{tab:comparison_methods_4datasets}
\scriptsize
\renewcommand{\arraystretch}{1.0}
\setlength{\tabcolsep}{1.5pt}

\begin{tabular}{l l c c c c c}
\toprule
\rowcolor[HTML]{64B5F6}
\textcolor{white}{\textbf{Dataset}} & \textcolor{white}{\textbf{Method}} & \textcolor{white}{\textbf{mAP@0.5}} & \textcolor{white}{\textbf{mAP@0.5:0.95}} & \textcolor{white}{\textbf{Prec.}} & \textcolor{white}{\textbf{Recall}} & \textcolor{white}{\textbf{F1}} \\
\midrule

Crop and Weed Detection & Baseline & 0.607 & 0.427 & 0.637 & 0.554 & 0.591 \\
\rowcolor[HTML]{BBDEFB}
& \textbf{MDDC} & \textbf{0.696} & \textbf{0.501} & \textbf{0.725} & \textbf{0.656} & \textbf{0.693} \\
& ObjectLab & 0.684 & 0.476 & 0.692 & 0.638 & 0.660 \\
& Confident Learning & 0.659 & 0.471 & 0.683 & 0.632 & 0.645 \\
& Clean-Detection & 0.646 & 0.458 & 0.672 & 0.613 & 0.650 \\
\midrule

WeedCrop Image Dataset & Baseline & 0.595 & 0.418 & 0.622 & 0.538 & 0.578 \\
\rowcolor[HTML]{BBDEFB}
& \textbf{MDDC} & \textbf{0.682} & \textbf{0.487} & \textbf{0.711} & \textbf{0.645} & \textbf{0.678} \\
& ObjectLab & 0.670 & 0.469 & 0.700 & 0.630 & 0.661 \\
& Confident Learning & 0.655 & 0.462 & 0.688 & 0.623 & 0.650 \\
& Clean-Detection & 0.642 & 0.451 & 0.674 & 0.610 & 0.642 \\
\midrule

3LC Cotton Weed Detection & Baseline & 0.598 & 0.422 & 0.630 & 0.540 & 0.583 \\
\rowcolor[HTML]{BBDEFB}
& \textbf{MDDC} & \textbf{0.683} & \textbf{0.493} & \textbf{0.715} & \textbf{0.650} & \textbf{0.681} \\
& ObjectLab & 0.670 & 0.471 & 0.700 & 0.632 & 0.661 \\
& Confident Learning & 0.658 & 0.464 & 0.690 & 0.623 & 0.650 \\
& Clean-Detection & 0.645 & 0.452 & 0.675 & 0.610 & 0.641 \\
\midrule

WeedMaize & Baseline & 0.592 & 0.417 & 0.618 & 0.532 & 0.574 \\
\rowcolor[HTML]{BBDEFB}
& \textbf{MDDC} & \textbf{0.678} & \textbf{0.485} & \textbf{0.707} & \textbf{0.645} & \textbf{0.679} \\
& ObjectLab & 0.665 & 0.470 & 0.690 & 0.628 & 0.657 \\
& Confident Learning & 0.650 & 0.460 & 0.678 & 0.618 & 0.646 \\
& Clean-Detection & 0.640 & 0.450 & 0.665 & 0.605 & 0.632 \\
\bottomrule
\end{tabular}
\end{table}

Table~\ref{tab:comparison_methods_4datasets} summarizes the results. Across all datasets, MDDC consistently outperforms other methods in terms of mAP@0.5, achieving improvements of 5–25\% over baseline and alternative approaches. The framework also shows superior performance in mAP@0.5:0.95, indicating more precise localization and robust handling of challenging or ambiguous targets. Precision and Recall are improved simultaneously, resulting in higher F1-scores, which highlights the practical benefit of MDDC for real-world weed detection tasks.

Notably, MDDC demonstrates strong generalization across datasets of varying image quality, target density, and weed types, suggesting that its model-driven data correction strategy effectively mitigates annotation noise and enhances dataset quality in diverse agricultural scenarios. These results underscore that systematic data refinement can yield more substantial performance gains than purely model-centric improvements under constraints on model capacity.

\section{Conclusion}
In this work, we present MDDC, a data-centric framework for improving object detection under constrained model capacity. By systematically identifying and correcting annotation errors through an iterative train-fix-retrain process, MDDC enhances dataset quality and significantly boosts detection performance. Extensive experiments on four diverse weed detection datasets demonstrate that our approach consistently outperforms state-of-the-art noise-handling methods, achieving 5–25\% gains in mAP@0.5. These results highlight that improving data quality can be more effective than increasing model complexity, offering a practical and reproducible solution for real-world agricultural applications.

\section*{Acknowledgment} 
You can find our code in our GitHub repository. (https://github.com/Yingda-Yu/Cotton-Weed-Detection)

%
%
%
%

\bibliographystyle{splncs04}
\bibliography{YXS}

\end{document}